\documentclass[letterpaper, 10 pt, conference]{ieeeconf}  % Comment this line out if you need a4paper

\IEEEoverridecommandlockouts                              % This command is only needed if 
\usepackage{graphicx}
\usepackage{multirow}
\usepackage{amsmath}
\usepackage{tabularx}
\usepackage{makecell}
\usepackage{booktabs}

\title{\LARGE \bf
The Price Is Not Right: Neuro-Symbolic Methods Outperform VLAs on Structured Long-Horizon Manipulation Tasks with Significantly Lower Energy Consumption
}

\author{Timothy Duggan$^{1}$, Pierrick Lorang$^{1,2}$, Hong Lu$^{1}$, and Matthias Scheutz$^{1,*}$%
  \thanks{*This work was in part supported by ONR grant \#N00014-24-1-2024.}%
  \thanks{\raggedright $^{1}$Human-Robot Interaction Lab, Tufts University, Medford, MA, USA.
  {\tt\small \{timothy.duggan, pierrick.lorang, hong.lu663424, matthias.scheutz\}@tufts.edu}}
  \thanks{\raggedright $^{2}$AIT Austrian Institute of Technology GmbH, Center for Vision, Automation \& Control, Vienna, Austria.
  {\tt\small pierrick.lorang@ait.ac.at}}%
}

\begin{document}

\maketitle
\thispagestyle{empty}
\pagestyle{empty}

%%%%%%%%%%%%%%%%%%%%%%%%%%%%%%%%%%%%%%%%%%%%%%%%%%%%%%%%%%%%%%%%%%%%%%%%%%%%%%%%
\begin{abstract}
  Vision-Language-Action (VLA) models have recently been proposed as a pathway toward generalist robotic policies capable of interpreting natural language and visual inputs to generate manipulation actions. However, their effectiveness and efficiency on structured, long-horizon manipulation tasks remain unclear. In this work, we present a head-to-head empirical comparison between a fine-tuned open-weight VLA model ($\pi_0$) and a neuro-symbolic architecture that combines PDDL-based symbolic planning with learned low-level control. 

  We evaluate both approaches on structured variants of the Towers of Hanoi manipulation task in simulation while measuring both task performance and energy consumption during training and execution. On the 3-block task, the neuro-symbolic model achieves 95\% success compared to 34\% for the best-performing VLA. The neuro-symbolic model also generalizes to an unseen 4-block variant (78\% success), whereas both VLAs fail to complete the task. During training, VLA fine-tuning consumes nearly two orders of magnitude more energy than the neuro-symbolic approach.
  
  These results highlight important trade-offs between end-to-end foundation-model approaches and structured reasoning architectures for long-horizon robotic manipulation, emphasizing the role of explicit symbolic structure in improving reliability, data efficiency, and energy efficiency.
  Code and models are available at \texttt{https://price-is-not-right.github.io}

\end{abstract}

%%%%%%%%%%%%%%%%%%%%%%%%%%%%%%%%%%%%%%%%%%%%%%%%%%%%%%%%%%%%%%%%%%%%%%%%%%%%%%%%

\section{Introduction}

% \begin{figure*}[t]
% \centering
% \includegraphics[width=\textwidth]{figures/overview_horizontal.png}
% \caption{Overview of the experimental comparison between the Vision-Language-Action (VLA) models and the neuro-symbolic model (NSM). Both approaches receive identical sensory inputs, but differ in how planning and control are structured. All training, inference, and simulation components run on the same hardware, and power/energy are measured at the system level.}
% \label{fig:overview}
% \end{figure*}

\begin{figure}[t]
\centering
\includegraphics[width=1\columnwidth]{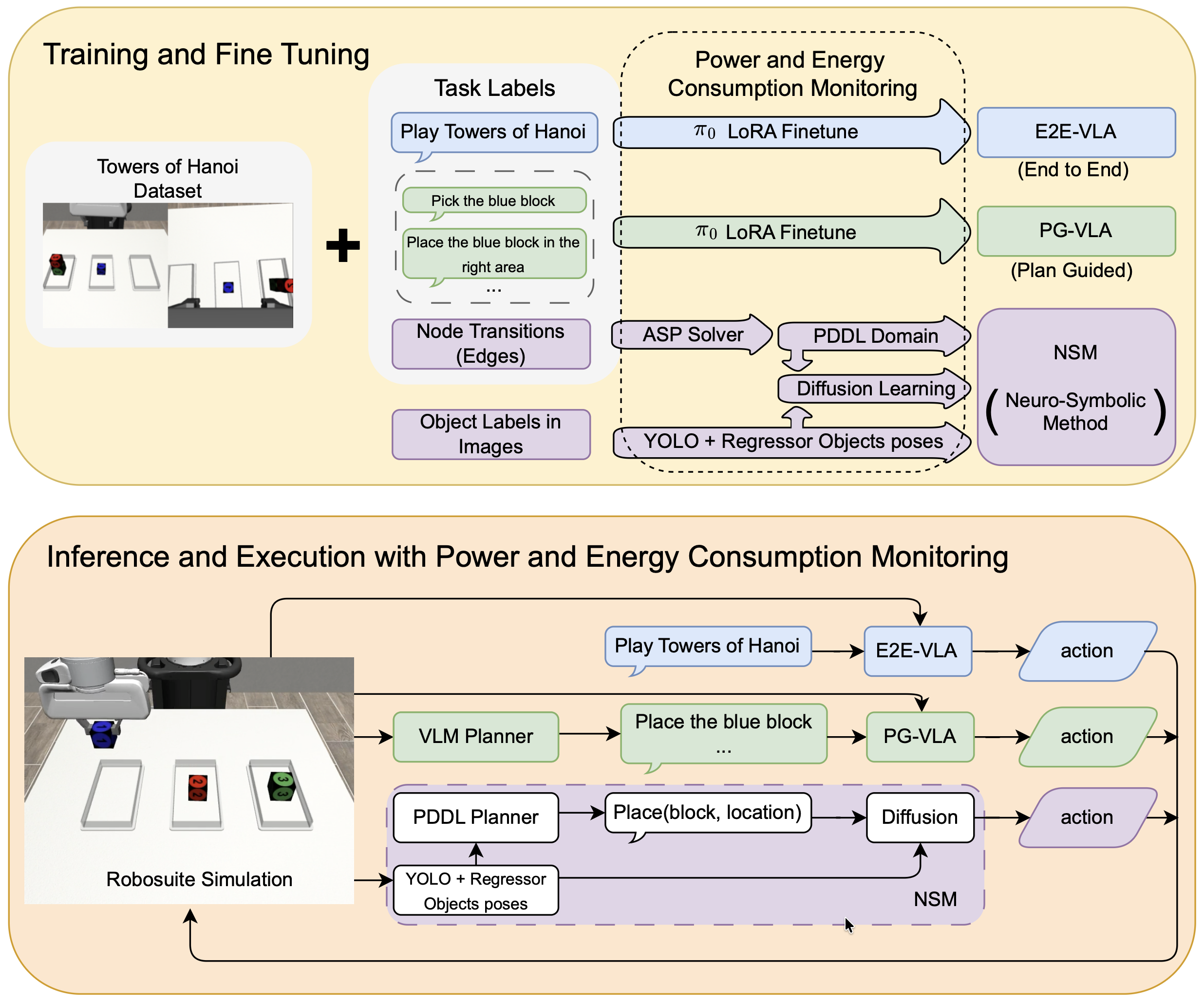}
\caption{Overview of the experimental comparison between VLA models and the NSM. Both receive identical sensory inputs from simulation. VLAs produce actions conditioned on either a high-level task description or planner guidance, while the NSM plans symbolically and executes via learned policies. Power and energy are monitored during training and inference.}
\label{fig:overview}
\end{figure}

Vision-Language-Action (VLA) models are gaining attention in robotics
due to their ability to interpret natural language task instructions
in conjunction with visual inputs to produce manipulation actions.
Such models are often viewed as a pathway toward generalist robotic
policies that can be adapted to a variety of tasks. However, their
suitability for structured, long-horizon manipulation problems
remains unclear, particularly when considering the substantial computational and
energy costs associated with fine-tuning and deployment.

To evaluate whether the use of VLAs is justified in this setting, we
conduct a head-to-head comparison between a state-of-the-art open-weight
VLA model, $\pi_0$~(OpenPi)~\cite{black2024pi0visionlanguageactionflowmodel}, and a recently published neuro-symbolic architecture that combines a high-level PDDL-based symbolic planner with diffusion-based low-level control~\cite{Lorang_Lu_Huemer_Zips_Scheutz_2025}. This
alternative model leverages the interpretability and structured
reasoning capabilities of symbolic methods alongside the adaptive and
continuous control advantages offered by neural network-based
approaches.  We designed a ``Towers of Hanoi''
manipulation task within the {\em Robosuite} simulation environment to benchmark multi-step robotic manipulation capabilities
comprehensively.  Both approaches are trained from programmatically generated demonstrations with the same observation and action modalities.

Importantly, we evaluate two VLA configurations: an end-to-end model and a planner-guided variant.
The end-to-end VLA is not provided with explicit task rules or symbolic structure; instead, it must infer relevant constraints from demonstration data alone.
The planner-guided VLA is fine-tuned to execute subtasks under the guidance of an external planner, allowing us to isolate execution performance from planning quality.

Our evaluation focuses on task performance, energy consumption, and
generalization to task variations. We record training time,
CPU/GPU utilization, and total energy usage for both approaches. The
results show that, in this structured long-horizon setting, the
neuro-symbolic model achieves substantially higher task success while
consuming significantly less energy than the VLA model. These
findings suggest that for structured manipulation tasks requiring
sequential reasoning, explicitly incorporating symbolic structure can
provide significant performance and efficiency advantages.

While this benchmark focuses on a structured manipulation domain, it provides a controlled setting for analyzing performance–efficiency trade-offs between architectural paradigms.

%%%%%%%%%%%%%%%%%%%%
% Related Works
%%%%%%%%%%%%%%%%%%%%
\section{Related Works}
\subsection{Vision-Language-Action Models}
Recent advancements in end-to-end robotics foundational models have
aimed at enabling robots to generalize across tasks, environments and
morphologies. VLA models, in particular, integrate vision, language,
and control capabilities through large scale pre-training on robot
demonstration data with the potential of providing generalist robot
policies~\cite{hu2024generalpurposerobotsfoundationmodels}. Similarly,
after rigorous evaluation of multitask robot manipulation policies,
recent findings suggest that multitask pretraining can improve the
success rate and robustness of policies while speeding up the learning
of new tasks~\cite{trilbmteam2025carefulexaminationlargebehavior}.

Several recent VLA models have explored distinct aspects of multi-modal
integration. Kim et al.\ introduce {\em OpenVLA}, a model designed for
generalized task planning and execution through large-scale training
on paired vision-action data
\cite{kim2024openvlaopensourcevisionlanguageactionmodel}. {\em UniVLA}
by Bu et al.\ expands on this by employing a unified architecture to
handle a broader range of tasks across multiple robot morphologies and
environments, emphasizing versatility and adaptability
\cite{bu2025univlalearningacttaskcentric}.
NVIDIA's {\em GR00T} model further advances these ideas by integrating
a dual system design where a vision-language module and a diffusion
transformer module are trained together for tight coupling between the
two~\cite{nvidia2025gr00tn1openfoundation}. Physical Intelligence's {\em $\pi_0$} focuses on
dexterous manipulation and has an action chunking architecture with
flow matching that enables high frequency control~\cite{black2024pi0visionlanguageactionflowmodel}.

However, questions remain regarding the generalization capabilities of current VLAs. Current VLA models frequently fail when faced with complex,
out-of-distribution, or multi-step
tasks~\cite{guruprasad2024benchmarkingvisionlanguage}. For instance,
Zhang et al. highlight that {\em OpenVLA} struggles with
generalization, skill transfer, and long-horizon
planning~\cite{zhang2024vlabenchlargescalebenchmarklanguageconditioned}. In addition to these performance limitations, current VLA research rarely reports detailed measurements of computational and energy costs associated with deployment, particularly on embedded robotic systems.

%%%%% Related Works Subsection %%%%%
\subsection{Neuro-symbolic Approaches for Task and Motion Planning}

Neuro-symbolic approaches have been widely studied as a way to
integrate high-level symbolic reasoning with low-level continuous
control, particularly in ``Task and Motion Planning'' (TAMP)
domains~\cite{steccanella2022state, karia2022relational, Kokel2021,
  guan2022leveraging, kumar2022learning, garrett2020online}. These
systems often rely on structured symbolic representations and action
operators, which can be manually specified or learned from
demonstrations~\cite{Konidaris_Kaelbling_Lozano,
  Manschitz_Kober_Gienger_Peters_2014, peorl-Yang,
  Pertsch_Lee_Wu_Lim_2021, Lorang_Goel_Shukla_Zips_Scheutz_2024},
enabling long-horizon behavior while maintaining interpretability and
modularity. Logic-based formalisms, such as {\em Answer Set
  Programming} (ASP), have also been used to extract symbolic models
from raw data, reducing supervision requirements though typically
assuming discrete state and action
abstractions~\cite{Bonet_Geffner_2020,DBLP:journals/corr/abs-2105-10830}.

Historically, these neuro-symbolic architectures demonstrate the
feasibility of structured, interpretable planning and control in data-efficient settings~\cite{Tanwani_Yan_Lee_Calinon_Goldberg_2021,
  Zhu_Stone_Zhu_2022, Teng_Chen_Ai_Zhou_Xuanyuan_Hu_2023,Kumar_McClinton_Chitnis_Silver_Lozano-Pérez_Kaelbling_2023,
  chi2024diffusionpolicyvisuomotorpolicy, Kokel2021,
  Lorang_Lu_Scheutz_2025}. These developments highlight how agents can
achieve long-horizon, modular, and interpretable behavior while
remaining data-efficient, providing a strong historical baseline for
analyzing the computational and energy trade-offs of modern VLAs.
Among the recent works, the approach by Lorang et
al.~\cite{Lorang_Lu_Huemer_Zips_Scheutz_2025} provides a practical
example of a neuro-symbolic system capable of co-learning both
symbolic domains and low-level policies from very few demonstrations.
This method illustrates how agents can plan and act efficiently in structured environments without relying on large-scale pretraining, making it a suitable baseline for comparison with VLAs in terms of computational and energy cost.

%%%%%%%%%%%%%%%%%%%%%%%%%%%%%%
% Preliminaries
%%%%%%%%%%%%%%%%%%%%%%%%%%%%%%
\subsection{Preliminaries: Neuro-Symbolic Planning-Diffusion Models}

\textbf{Symbolic Planning.} Symbolic planning builds upon a formal
domain description $\sigma = \langle \mathcal{E}, \mathcal{F},
\mathcal{S}, \mathcal{O}\rangle$, where $\mathcal{E}$ is a set of
entities, $\mathcal{F}$ a set of boolean or numerical predicates over
entities, $\mathcal{S}$ a set of symbolic states formed by grounded
predicates, and $\mathcal{O}$ a set of operators. Each operator $o \in
\mathcal{O}$ is defined by preconditions $\psi$ and effects $\omega$
over predicates. A grounded operator $\hat{o}$ binds objects to
parameters and can be applied if its preconditions hold, updating the
state according to its effects. A planning task $T=(\mathcal{E},
\mathcal{F}, \mathcal{O}, s_0, s_g)$ seeks a plan
$\mathcal{P}=[o_1,\ldots,o_{|\mathcal{P}|}]$ that transitions from
initial state $s_0$ to goal state $s_g$~\cite{mcdermott_pddl_1998}.

\textbf{Diffusion Imitation Learning (IL).}  Imitation learning (IL)
aims to learn a policy $\pi(\tilde{s})$ from expert demonstrations
$\{(\tilde{s}_t, a_t, \tilde{s}_{t+1})\}_{t=0}^{T}$, where
$\tilde{s}_t$ denotes a continuous state, $a_t$ the expert action, and
$\tilde{s}_{t+1}$ the resulting state. The policy is trained by
minimizing the mean squared error between predicted and expert
actions: $\mathcal{L}(\pi) = \frac{1}{T} \sum_{t=0}^{T} \|
\pi(\tilde{s}_t) - a_t \|^2 .$ Unlike reinforcement learning, IL
bypasses exploration and reward engineering, providing a more
data-efficient way to acquire complex behaviors directly from
demonstrations.

Diffusion policies~\cite{chi2023diffusionpolicy} are imitation
learning methods for continuous control that adapt diffusion models
from generative modeling to policy learning. Expert actions are
perturbed with Gaussian noise during training, and a denoising network
$\epsilon_{\theta}$ is optimized to recover the original actions
conditioned on the state. At inference, the learned reverse process
iteratively refines noisy samples to generate actions, yielding a
stochastic policy $p_{\theta}(a_t \mid s_t)$.

\textbf{Neuro-Symbolic Architecture.} Neuro-symbolic architectures
combine symbolic reasoning with neural control. A planner solves a
STRIPS task $T = \langle \mathcal{E}, \mathcal{F}, \mathcal{O}, s_0,
s_g \rangle$ to produce a plan
$\mathcal{P}=[o_1,\ldots,o_{|\mathcal{P}|}]$, where each operator
$o_i$ is refined into a neural skill $\pi_i \in \Pi$. Each skill
$\pi_i$ interacts with the environment to realize the operator's
effects $\omega_i$, transitioning the system from a state $s$ to a new
state $s'$. This layered approach enables flexible execution in
continuous spaces while maintaining high-level task abstraction.

\section{Comparing State-of-the-Art VLA and Neuro-Symbolic Models}

\subsection{Fine-tuning the Vision Language Action Model} 

$\pi_0$ was selected as a representative open-weight VLA model with publicly available fine-tuning infrastructure. 
Two separate checkpoints were created using LoRA fine-tuning: one End-to-End (E2E-VLA) and one Planner-Guided (PG-VLA).
More details on each can be found in Sec.~\ref{baselines}. LoRA fine-tunings were performed for 30k steps on our custom dataset using the default hyperparameters provided in the official OpenPi fine-tuning scripts to ensure reproducibility.~\footnote{At the time of our experiments, the official OpenPi fine-tuning pipeline included a delta-action transformation enabled by default in the released training scripts. This transform is intended for datasets containing absolute action targets. Although our dataset already consisted of relative (delta) actions, we retained the default configuration to remain consistent with the provided checkpoint and training code. All reported VLA results reflect this setting. Subsequent updates to the OpenPi codebase have modified this default behavior. We verified that disabling this transform does not qualitatively alter the performance trends reported in this work.}
PaliGemma 2B LoRA was chosen as the
vision-language backbone and Gemma 300M LoRA was chosen for the action
header. The exact configuration can be seen in our repository.

\subsection{Inference with the Vision Language Action Model} 
Inference was performed inside the docker container provided by
$\pi_0$. Robosuite was added to the container to access our custom
environment. For both E2E-VLA and PG-VLA, progress is tracked using
detectors in the simulation. E2E-VLA's command, ``Play Towers of
Hanoi'', remains constant throughout the episodes. PG-VLA receives its
next command after successfully completing its current command. If
the VLA does not progress to the next sub-task within 750 steps, the
episode is terminated.

\subsection{Planning}

\subsubsection{VLA latent space}
E2E-VLA uses its latent space planning to determine the moves it takes.

\subsubsection{VLM}
Since the $\pi_0$ team uses VLMs to break high-level tasks down into
low-level language commands, we evaluate three VLMs on their ability to
generate plans as sequences of pick-and-place language commands for
the Towers of Hanoi tasks. The three VLMs evaluated are GPT-5, Qwen,
and PaliGemma
\cite{openai2025gpt-5,bai2025qwen25vltechnicalreport,beyer2024paligemmaversatile3bvlm}. We
access GPT-5 through the OpenAI API. We download and evaluate Qwen
and PaliGemma locally on a GPU machine. The Qwen variant used is
Qwen2.5-7b-VL-Instruct. This variant has 7 billion parameters and has
been trained to produce output in the instructed format. The
PaliGemma variant used is PaliGemma-3b-mix-448 variant. This variant
has 3 billion parameters and has been trained on a variety of vision
language tasks. Each VLM takes in an initial image of the Towers of
Hanoi task, a goal image, and a prompt explaining the environment and
the rules of Towers of Hanoi. The VLMs are prompted to output a
sequence of pick or place natural language commands separated by the
new line character. To isolate execution performance from planning quality, the planner-guided VLA is evaluated using valid, optimal plans.

\subsection{Neuro-symbolic Architecture}

We utilize prior work demonstrating that neuro-symbolic models can be
learned from few
demonstrations~\cite{Lorang_Lu_Huemer_Zips_Scheutz_2025} to construct
our neuro-symbolic model (NSM) which combines PDDL-based symbolic
planning with low-level controls. When performing the task, the model
first generates a high-level plan using the symbolic planner, detects the object pose and then
executes each step by using a trained diffusion policy. All diffusion policies operate on relative pose observations--object poses expressed
with respect to the end effector--and output end-effector displacement
actions. To keep the input modalities the same as those of VLAs, the
NSM receives only images and proprioceptive information. Both the
high-level planning domain and the low-level controls are learned
directly from demonstrations. The NSM learns symbolic abstractions and low-level policies from demonstrations without manually specified task rules or pre-defined action schemas.

\textbf{Symbolic Abstraction.} From raw demonstration trajectories
$\mathcal{D}$, we extract node transitions $\tau^{node} = (n, l, n')$,
where $n$ and $n'$ are high-level states and $l$ is a human-assigned
label. These transitions form a graph $G=\langle V,E,L\rangle$ whose
nodes represent abstract states and edges represent skills.  To
compact the structure, we compute a minimal bisimulation $\bar{G}$,
removing redundant states while preserving equivalence.  Using an
ASP-based solver~\cite{Bonet_Geffner_2020, kr2021rbrg}, we infer a
symbolic domain $\sigma = \langle \mathcal{E}, \mathcal{F},
\mathcal{S}, \mathcal{O}\rangle$ in PDDL form.  This yields a symbolic
abstraction of the demonstrations suitable for classical planning.

\textbf{Object detection.} To ensure comparability with VLA inputs,
the NSM also takes as input images and proprioceptive data rather than
object-space information, which was used in prior
work~\cite{Lorang_Lu_Huemer_Zips_Scheutz_2025}.  Specifically, we
train a YOLOv8~\cite{Jocher_Ultralytics_YOLO_2023} bounding-box
detector combined with a lightweight gradient boosting regressor to
estimate 3D object poses from two camera views: a static camera and a
wrist-mounted camera on the end effector. These are the same two cameras used by the VLAs.  These estimate poses, combined with
proprioceptive data, are then used to compute relative transformations
between the end effector and detected objects and fed as inputs to the
control policies.

\textbf{Training Controls.} Each operator $o_i \in \mathcal{O}$ is
associated with a neural policy $\pi_i$ trained from its demonstration
segments.  Inspired by the options framework~\cite{SUTTON1999181}, we
further decompose each skill into action-step sub-policies $\pi_{i,j}$
with termination conditions.  To ensure sample efficiency, we filter
observations through a task-relevant feature selector $\phi$ that selects
only operator-relevant objects $\mathcal{E}_{o_i}$ and expresses them
in coordinates relative to the end effector.  Policies are trained
using diffusion models~\cite{chi2023diffusionpolicy}, which capture
multi-modal action distributions and improve robustness.

\textbf{Execution of the Framework.} At test time, the user specifies
an initial and goal state, which are mapped into a PDDL planning
instance $T=\langle
\mathcal{E},\mathcal{F},\mathcal{O},s_0,s_g\rangle$.  A classical
planner (MetricFF~\cite{hoffmann2003metric}) computes an abstract plan
$\mathcal{P}=[o_1,\dots,o_{|\mathcal{P}|}]$.  Each operator $o_i$ is
realized by invoking its policy $\pi_i$, which internally sequences
action-step controllers $\pi_{i,j}$ until their learned termination
conditions are met. This hierarchical composition enables
generalization to variations within the task domain, solving
long-horizon TAMP problems from few demonstrations.

%%%%%%%%%%%%%%%%%%%%%%%%%%%%%%
% Methods
%%%%%%%%%%%%%%%%%%%%%%%%%%%%%%
\section{Evaluation Methodology}

For the model comparison, we were particularly interested in
determining performance under both task-based and energy-based measures.
Task-based measures include successful task completion rates as well
as generalization to novel task variations not encountered during
training.  Energy-based measures include energy expenditure during
fine-tuning and training, as well as during inference.  In addition, we evaluated three VLMs which were used to break down the problem into atomic pick-and-place actions for the VLA on their energy expenditure and plan correctness.

\subsection{Hardware}
All fine-tuning, training, inference, and experiments were performed on
a single NVIDIA GeForce RTX 4090 with 24GB of memory. GPU Power
consumption metrics were measured using Weights \&
Biases~\cite{wandb}. CPU power consumption was measured by accessing the RAPL logs.

Idle GPU power averaged approximately 25~W and idle CPU power
approximately 2.5~W. Idle power consumption was included in all
reported measurements to reflect total system-level energy usage
during training and inference.

Power usage was logged periodically during execution and training.
Due to variability in logging intervals imposed by the monitoring
framework, reported energy values were computed by numerically
integrating measured power over recorded timestamps.

\subsection{Simulation and Experimental Environment}
All simulations were performed in {\em Robosuite} in the tabletop
environment designed for the robot to perform a modified version of
{\em Towers of Hanoi} as seen in Fig.~\ref{fig:overview}. We used
blocks instead of discs, and rectangular areas instead of rods to
reduce manipulation failures.  The blocks are numbered; higher numbers
indicated ``larger blocks''. Blocks have different sizes, as well as
different colors per block size. The experiments are performed in
three different task environments:
\begin{enumerate}
    \item \textbf{Individual Move:} A basic pick and place task where the
      agent must place a single block onto another designated
      block or area. This serves as the simplest benchmark for object
      manipulation.
    \item \textbf{Three-block Towers of Hanoi:} Each episode begins
      with a tower of three blocks stacked on the left platform. The
      goal is to reconstruct the tower on the right platform while
      respecting the constraint that a larger block can never be
      placed on top of a smaller one.
    \item \textbf{Four-block Towers of Hanoi:} The most challenging
      variant, where the task starts with all four blocks stacked on
      the left platform. The agent must transfer the complete tower to
      the right platform under the same size-ordering constraint.
\end{enumerate}

The robot platform used in our experiments is the \textit{Franka
  Panda}.  We selected this robot because the $\pi_0$ model, which we
employ as a baseline, was originally trained and fine-tuned on the
LIBERO simulation benchmark using the same platform.  This choice
ensures that our evaluation environments remain closely aligned with
those used to train the $\pi_0$ checkpoint, even prior to its
fine-tuning on our specific tasks.

\begin{figure}[t]
\centering
\vspace{0.2cm}
\includegraphics[width=0.8\columnwidth]{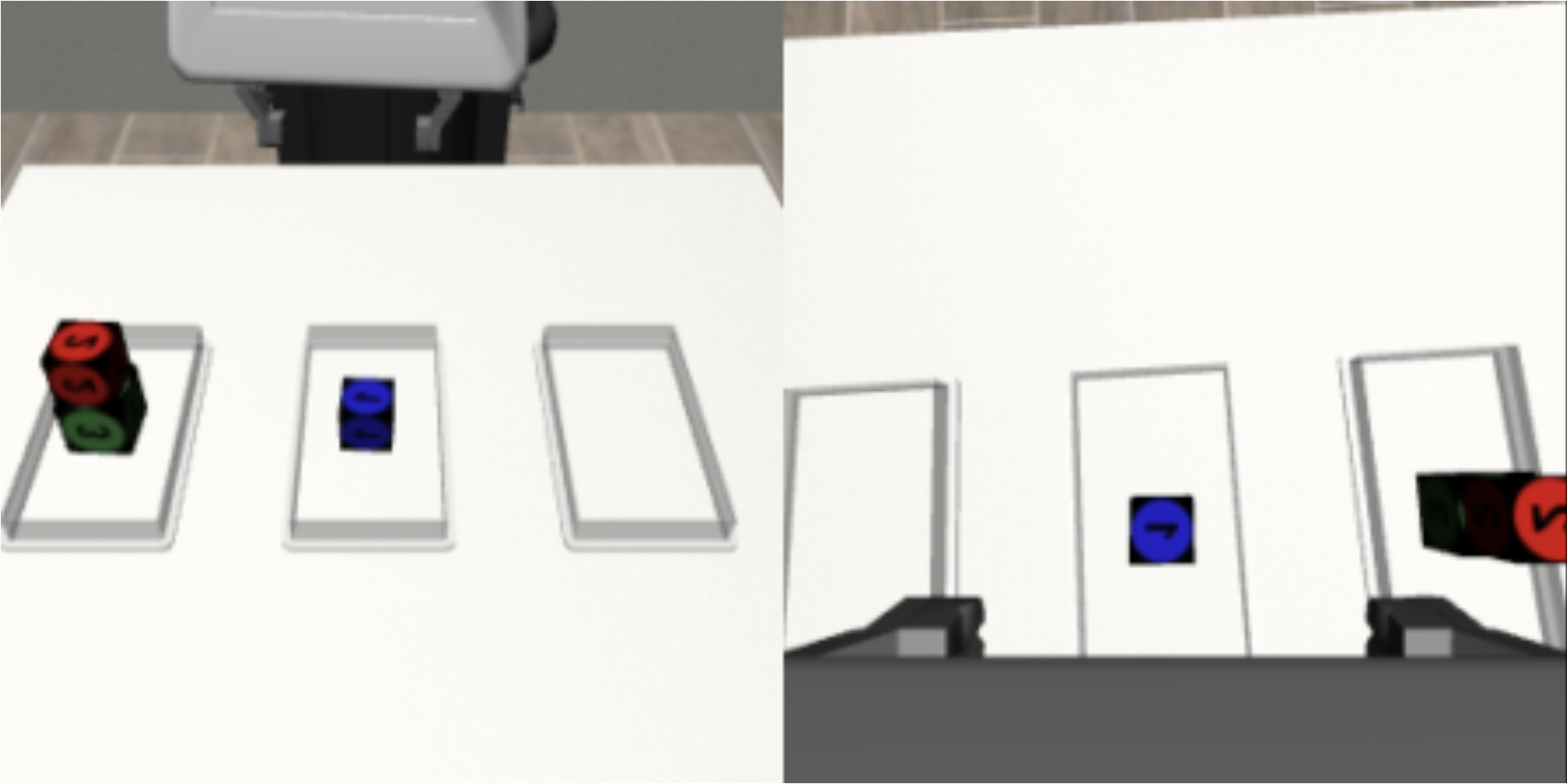}
\caption{Example observations from the dataset. Left: Agent-view RGB image. Right: Wrist-mounted camera RGB image.}
\label{fig:agent_and_wrist_images}
\end{figure}

\subsection{Training Data} \label{sec:training_data}

Training data was automatically generated using a Python script that
controlled the robot using scripted grasp-and-place primitives to complete Towers
of Hanoi tasks.

\textbf{VLA Training Dataset.} We collected a dataset of $300$
episodes, which can be found on the paper's website. The first $150$
episodes consist of full Towers of Hanoi runs with a random selection
of $3$ out of the $4$ possible blocks. The remaining $150$ episodes
begin from random valid configurations of blocks placed on different
platforms, while respecting the rules of the task. In both cases,
block positions on the platforms were varied in the $x,y$ plane
according to a Gaussian distribution with a standard deviation of $1$
cm.

\textbf{NSM Training Dataset}. The NSM training data was substantially smaller: only $50$ demonstrations of randomly sampled
\textit{Stacking} tasks (pick and place pairs) were used. Importantly, the NSM never directly
observes a Towers of Hanoi resolution during demonstrations; instead,
it infers the rules and symbolic planning domain from these simpler
stacking demonstrations. Each demonstration consists of (images,
proprioception, action) sequences, along with a high-level linkage of
nodes between demonstrations
(see~\cite{Lorang_Lu_Huemer_Zips_Scheutz_2025} for technical
details). We used the same random sampling as for the VLA training
data.

Although the VLA is trained on full task trajectories while the NSM is trained only on stacking demonstrations, both datasets share identical sensory and action formats. The primary methodological difference lies in how each architecture leverages its training data: the NSM explicitly abstracts symbolic structure from partial demonstrations, whereas the VLA must implicitly infer such structure from complete trajectories.

Here is the overall demonstration data we provide to all agents.
\begin{itemize}
    \item \texttt{image (256,256,3)}: RGB image from the Robosuite
      agentview camera. See Fig.~\ref{fig:agent_and_wrist_images}
    \item \texttt{wrist\_image (256,256,3)}: RGB image from the
      Panda's wrist camera. See Fig.~\ref{fig:agent_and_wrist_images}
    \item \texttt{state (8,)}: The grippers position, orientation
      (axis angle), and finger positions.
    \item \texttt{actions (7,)}: Robosuite OSC\_POSE controller action
      (dx, dy, dz, droll, dpitch, dyaw, gripper)
    \item \texttt{task (string)}: The natural language command (VLA specific)
    \item \texttt{task (node transitions)}: Feasible transitions between \textit{Stacking} tasks (NSM specific)  
\end{itemize}

The following shows the \textbf{Natural Language Sub-Task Formats}
present in the training dataset:
\begin{itemize}
    \item \textbf{Pick} the [\textit{blue} / \textit{red} / \textit{green} / \textit{yellow}] block.
    \item \textbf{Place} the [\textit{blue} / \textit{red} / \textit{green} / \textit{yellow}] block in the [\textit{left} / \textit{middle} / \textit{right}] area.
    \item \textbf{Place} the [\textit{blue} / \textit{red} / \textit{green} / \textit{yellow}] block on the [\textit{blue} / \textit{red} / \textit{green} / \textit{yellow}] block.
\end{itemize}

\subsection{Baselines}\label{baselines}

We evaluated the following models:

\textbf{End-to-End VLA (E2E-VLA):} A vision-language-action model
fine-tuned on the training datasets using a single high-level command,
``Play Towers of Hanoi'', applied at every frame of the training
episodes. This model learns the task as a monolithic sequence without
explicit decomposition into sub-tasks and does not use any external
planning mechanism.
    
\textbf{Planner-Guided VLA (PG-VLA):} A vision-language-action model
fine-tuned on our datasets using structured sub-task commands (see
Sec.~\ref{sec:training_data}), in combination with an external
planner. This approach decomposes the overall task into smaller
natural language sub-goals, enabling stepwise guidance during
execution. We separate the planning performance from the acting
performance evaluations, and consider that PG-VLA always has access to the optimal plan to isolate execution performance from planning quality.
    
\textbf{Neuro-Symbolic Model (NSM):} A baseline method that learns
both high-level symbolic planning and low-level continuous control
directly from
demonstrations~\cite{Lorang_Lu_Huemer_Zips_Scheutz_2025}. The NSM
extracts abstract operators and builds a PDDL domain from a small
number of demonstrations, executing the task by sequencing learned
neural policies according to the planned operator sequence.

\subsection{Metrics for Comparison}
We evaluated the performance of all agents on both short-horizon
(\textit{Individual Moves}) and long-horizon tasks (\textit{3-block and
  4-block Towers of Hanoi}). For planning, we computed the number of
optimal, suboptimal, and invalid plans over 50 episodes. For the
\textit{E2E-VLA} agent, planning performance was inferred indirectly
by observing task advancement relative to the optimal plan. In
contrast, for the \textit{PG-VLA} agent and the \textit{NSM} agent, we
directly analyzed the output plans produced by the VLM-based planner
and the classical PDDL planner, respectively. Task progress was
monitored using a detector function implemented solely for evaluation;
agents had no access to this detector. Runs exceeding 750 time
steps--approximately double the maximum number of steps observed in successful training trajectories--were considered failures and terminated. Overall
success rate, incorporating both planning and execution, was used as
the primary performance metric. The \textit{Individual Move} task evaluates a single pick-and-place action extracted from the 3-block Towers of Hanoi sequence, isolating low-level execution performance independent of long-horizon planning.

To assess both utility and efficiency, we additionally recorded
energy-based metrics. For model training and fine-tuning, we compared
total GPU energy consumption, mean GPU power, mean CPU utilization,
and overall training time. During execution and inference, we measured
mean GPU and CPU utilization per task, mean execution episode
duration, and task progression rates (e.g., fraction of sub-tasks
successfully completed at each step). These metrics allow evaluating
not only task success but also the computational and energy cost of
each approach.

%%%%%%%%%%%%%%%%%%%%%%%%%%%%%%
% Results
%%%%%%%%%%%%%%%%%%%%%%%%%%%%%%
\section{Results}

\subsection{Energy Consumption}
\subsubsection{Fine-tuning \& Training}
Table~\ref{tab:training_fine-tuning} reports the power and energy
consumption and training times of our models.  Each VLA LoRA
fine-tune took over 1.5 days to complete whereas the NSM
completed training in 34 minutes.  The VLA LoRA fine-tunes also had
about 50~W higher average power consumption than NSM
training. The VLA LoRA fine-tunes consumed nearly two orders of magnitude more energy than NSM training.

\begin{table}[t]
    \vspace{0.2cm}
    \centering
    \renewcommand{\arraystretch}{1.2}
    \caption{Training hardware metrics comparing VLA LoRA fine-tuning (E2E-VLA, PG-VLA) and NSM training. Best values in each row are bolded.}
    \begin{footnotesize}
        \begin{tabular}{lrrr}
            \toprule
            \textbf{Metric} & \textbf{E2E-VLA} & \textbf{PG-VLA} & \textbf{NSM} \\
            \midrule
            
            Time & 1d 16h 26m & 1d 15h 42m & \textbf{34m} \\
            
            \midrule
            \multicolumn{4}{l}{\textit{GPU Metrics}} \\
            Mean Util. (\%) & 100 & 100 & 100 \\
            Mean Power (W)  & 423.6 & 409.1 & \textbf{316.5} \\
            Energy (MJ)     & 61.7 & 58.5 & \textbf{0.65} \\
            
            \midrule
            \multicolumn{4}{l}{\textit{CPU Metrics}} \\
            Mean Util. (\%) & 3.12 & 3.13 & 10.5 \\
            Mean Power (W)  & 46.6 & \textbf{44.7} & 97.7 \\
            Energy (MJ)     & 6.8 & 6.4 & \textbf{0.2} \\
            
            \midrule
            \textbf{Total Energy (MJ)} & 68.5 & 64.9 & \textbf{0.85} \\
            
            \bottomrule
        \end{tabular}
    \end{footnotesize}
    \label{tab:training_fine-tuning}
\end{table}

\subsubsection{Inference \& Execution}
Table~\ref{tab:merged_results} reports power and energy consumption during evaluation. Both E2E-VLA and PG-VLA require GPU-backed inference, whereas the NSM does not. Although VLAs use roughly twice the CPU power of the NSM, their total power exceeds $5\times$ due to GPU usage.

Across all tasks except 4-block Towers of Hanoi, VLA episode energy is approximately an order of magnitude higher than that of the NSM. The 4-block case is an exception because episodes terminate early under the 750-step progress threshold, resulting in shorter runtimes and reduced accumulated energy.

Despite total power being only about $5\times$ greater, VLAs also require roughly $2\times$ longer per episode on individual block stacking tasks. Since energy scales with both power and time, this longer execution further amplifies the per-episode energy gap.

Episodes are terminated if no progress is made within 750 steps. Consequently, less successful models—particularly the VLA variants—often have shorter episode durations, which partially influences the observed differences in per-episode energy between E2E-VLA and PG-VLA.

\begin{table}[!t]
\centering
\vspace*{0.2cm}
\caption{Power, energy consumption, and task performance for Towers of Hanoi experiments. All values are averaged over 50 evaluation episodes. NSM does not use a GPU.}
\renewcommand{\arraystretch}{1.15}
\begin{footnotesize}
\begin{tabular}{c l r r r}
\toprule
\textbf{Setting} & \textbf{Metric} & \textbf{E2E-VLA} & \textbf{PG-VLA} & \textbf{NSM} \\
\midrule

\multirow{3}{*}{All Tasks}
 & GPU Power (W)   & 72.4  & 70.8  & \textbf{0} \\
 & CPU Power (W)   & 42.8  & 43.2  & \textbf{19.4} \\
 & Total Power (W) & 115.2 & 114.0 & \textbf{19.4} \\
\midrule

\multirow{3}{*}{\makecell{Individual\\Move}}
 & Success (\%)   & 87.0 & 59.6 & \textbf{99.0} \\
 & Duration (s)   & 13.8 & 12.4 & \textbf{6.3} \\
 & Energy (kJ)    & 1.59 & 1.41 & \textbf{0.12} \\
\midrule

\multirow{3}{*}{\makecell{3-Block\\Hanoi}}
 & Success (\%)        & 34.0 & 0.0  & \textbf{95.0} \\
 & Advancement (\%)    & 49.6 & 23.9 & \textbf{97.3} \\
 & Episode Energy (kJ) & 7.96 & 6.94 & \textbf{0.83} \\
\midrule

\multirow{3}{*}{\makecell{4-Block\\Hanoi}}
 & Success (\%)        & 0.0  & 0.0  & \textbf{78.0} \\
 & Advancement (\%)    & 2.5  & 3.6  & \textbf{84.4} \\
 & Episode Energy (kJ) & 5.77 & 4.96 & \textbf{1.44} \\
\bottomrule
\end{tabular}
\end{footnotesize}
\label{tab:merged_results}
\end{table}

\subsection{Task Performance}

\subsubsection{Individual Moves}
The NSM achieved an almost perfect success rate of 99\% in this task
of picking and placing a single block. The E2E-VLA had a 87.0\%
success rate and the PG-VLA had a 59.6\% success rate. We will discuss
the disparity between the task performance of the E2E-VLA and the
PG-VLA later. The NSM also completed individual move tasks in
about half the time of the E2E-VLA. The E2E-VLA had even worse speed
performance with a mean duration of 13.8 seconds.

\subsubsection{3-Block Towers of Hanoi}
For the 3-block version of Towers of Hanoi, the NSM achieved a success
rate of 95\%. The E2E-VLA completed 34\% of
the games and the PG-VLA did not complete a single game. We also look
at Task Advancement Rate which is the mean percentage of the task a
model completed. The NSM had a task advancement rate of the
97.3\%. The E2E-VLA had a Task Advancement Rate of 49.6\% with a
bimodal distribution of task advancement. Many episodes failed early
having only succeeded in the first 4 of 14 sub-tasks. Generally,
episodes that made it past sub-task 4 tended to complete the whole
task, with only a few exceptions.  The PG-VLA had a task success rate
of 0\% and an advancement rate of 23.9\%. Many of the episodes failed on sub-task 3 or 5 of
14. The PG-VLA's farthest progression was sub-task 9 of 14. None achieved
sub-task 10.  The low task performance rates of the VLA models are
particularly noteworthy because the evaluation configuration was included in their training data, hence one would expect to see a high performance. 

\subsubsection{4-Block Towers of Hanoi}
For the 4-block version of Towers of Hanoi which was not part of the
training data, the NSM achieved a success rate of 78\% and a task
advancement rate of 84.4\%. Both VLAs failed to complete a single
game. Their task advancement rates were also very low. Both VLAs
appeared to execute the trajectory for the 3-block Towers of Hanoi
game which is not too surprising in the case of the E2E-VLA because it
has never seen a 4-block Towers of Hanoi in its training data and
cannot generalize on its own. However, PG-VLA was given the proper
instructions and still was not able to execute them. The models were
supposed to pick the top block and place it on the middle platform. In
the 3-block Towers of Hanoi task, the first move is to place that block on
the right platform. The only successful instances of sub-task 2 occurred when the block was inadvertently released while passing over the middle platform.

\subsubsection{VLM Planners}
We queried each VLM with 10 pairs of initial and goal images per
Towers of Hanoi configuration. The results reported in
Table~\ref{tab:vlm_planner_comparison} summarize the 50 plans produced
for the five tower configurations of Hanoi. The GPU and CPU usages in
Table~\ref{tab:vlm_gpu_cpu_usages} are averaged across the 50
queries. Note that GPT-5 GPU and CPU stats were not accessible as GPT-5
is not open source. GPT-5 far outperformed Qwen and PaliGemma in
producing valid and optimal plans. However, since GPT-5 is much larger
than the other two models, we expect that its energy expenditure is substantially larger. These results indicate that VLMs are not
reliable and energy-efficient planners. This is consistent with Kambhampati's finding that large language models cannot reliably plan~\cite{kambhampati2024llmscantplanhelp}

\begin{table}[!t]
\centering
\vspace*{0.2cm}
\caption{Planning accuracy of VLM-based planners over 50 evaluation tasks.}
\renewcommand{\arraystretch}{1.2}
\begin{footnotesize}
\begin{tabular}{l r r r}
\toprule
\textbf{Metric} & \textbf{GPT-5} & \textbf{Qwen (7B)} & \textbf{PaLI-Gemma (3B)} \\
\midrule
Optimal (\%)    & \textbf{84} & 0   & 0   \\
Suboptimal (\%) & 0           & 0   & 0   \\
Invalid (\%)    & \textbf{16} & 100 & 100 \\
\bottomrule
\end{tabular}
\end{footnotesize}
\label{tab:vlm_planner_comparison}
\end{table}

\begin{table}[t]
\centering
\vspace*{0.2cm}
\caption{Per-query latency and hardware usage for VLM-based planners (mean over 50 queries). GPU/CPU statistics are unavailable for GPT-5 (API). Lower is better; best in bold.}
\renewcommand{\arraystretch}{1.15}
\begin{footnotesize}
\begin{tabular}{l r r r}
\toprule
\textbf{Metric} & \textbf{GPT-5} & \textbf{Qwen (7B)} & \textbf{PaLI-Gemma (3B)} \\
\midrule
Latency (s) & 63.1 & 1.83 & \textbf{0.22} \\
\midrule
\multicolumn{4}{l}{\textit{GPU metrics per query}} \\
Util. (\%)   & -- & 14.3 & 4.65 \\
Power (W)    & -- & 105.8 & \textbf{92.9} \\
Energy (J)   & -- & 193.6 & \textbf{20.4} \\
\midrule
\multicolumn{4}{l}{\textit{CPU metrics per query}} \\
Util. (\%)   & -- & 5.00 & 5.48 \\
Power (W)    & -- & 41.3 & \textbf{38.0} \\
Energy (J)   & -- & 75.6 & \textbf{8.4} \\
\midrule
Total Energy (J) & -- & 269.2 & \textbf{28.8} \\
\bottomrule
\end{tabular}
\end{footnotesize}
\label{tab:vlm_gpu_cpu_usages}
\end{table}

Overall, VLAs draw substantially higher power due to GPU usage, and longer episode durations further increase per-episode energy. We next examine whether this added cost yields improved task performance.

%%%%%%%%%%%%%%%%%%%%%%%%%%%%%%
% Discussion
%%%%%%%%%%%%%%%%%%%%%%%%%%%%%%
\section{Discussion}

The neuro-symbolic model outperforms the fine-tuned VLAs on the Towers
of Hanoi task in both task success and energy consumption. It also
generalizes to the 4-block variant despite not being trained on it,
demonstrating robustness on structured, multi-step manipulation.

The poor performance of PG-VLA in the 3-block task, even compared to
E2E-VLA, was unexpected. In the original $\pi_0$ work
\cite{black2024pi0visionlanguageactionflowmodel}, VLM-based task
decomposition improved long-horizon execution. We verified that the
PG-VLA was provided valid plans using commands from its training data.
One possible explanation is that the diversity within each command
category reduced execution fidelity, whereas the E2E-VLA partially
memorized the full 3-block trajectory. While generalization to the
4-block task was not expected for E2E-VLA, PG-VLA also failed to
generalize despite receiving correct instructions.

The VLA success rate of 34\% on the 3-block task highlights persistent
challenges for end-to-end models on structured long-horizon problems.
In our experiments, failures were primarily due to low-level execution
errors rather than complete breakdowns in task sequencing. The VLA
often retried missed grasps but repeatedly failed to accurately reach
target poses, suggesting sensitivity to trajectory quality and
variation within the training data. In particular, diversity
introduced through randomized block substitutions and spatial
perturbations may have reduced the model’s ability to execute precise
movements reliably.

Although VLA performance depends on data and hyperparameters, closing
the observed performance gap would likely still leave substantial
energy and training-time differences. The NSM was trained on fewer
demonstrations (50 stacking tasks) than the VLA (300 Hanoi games) yet
achieved near-perfect performance and better generalization through
explicit domain structure.

Overall, the results highlight important architectural trade-offs for
structured long-horizon tasks. While VLAs provide flexible end-to-end
learning, they remain sensitive to trajectory quality and incur
significant computational overhead during fine-tuning and inference.
For domains with explicit procedural constraints, such as industrial
assembly or rule-based manipulation, neuro-symbolic architectures that explicitly
represent task structure may provide practical advantages in
reliability, data efficiency, and energy consumption.

While this comparison focuses on a structured benchmark, the
computational demands of VLAs during fine-tuning and execution remain
substantial and may limit their practicality in many robotic settings. Because VLA policies require GPU-backed inference, their per-episode energy cost compounds under repeated deployment, potentially amplifying the practical gap observed in our experiments.  These findings suggest that continued investigation of hybrid
neuro-symbolic architectures remains well motivated.

%%%%%%%%%%%%%%%%%%%%%%%%%%%%%%
% Conclusion
%%%%%%%%%%%%%%%%%%%%%%%%%%%%%%
\section{Conclusion}

We presented a comprehensive empirical comparison between fine-tuned
Vision-Language-Action models and a neuro-symbolic architecture on a
structured long-horizon robotic manipulation task. In the 3-block
Towers of Hanoi benchmark, the neuro-symbolic approach achieved a
95\% success rate compared to 34\% for the best-performing VLA, while
consuming nearly two orders of magnitude less energy during training.
The neuro-symbolic model also generalized to the unseen 4-block
variant (78\% success), whereas both VLAs failed to complete a single
episode.

These results highlight important trade-offs between end-to-end
foundation-model approaches and structured reasoning architectures.
For manipulation tasks governed by explicit procedural constraints,
incorporating symbolic structure can yield substantial advantages in
reliability, data efficiency, and energy consumption.

We hope this work encourages more careful consideration of
performance–efficiency trade-offs when selecting architectures for
robotic manipulation.

%\appendix

\bibliographystyle{IEEEtran}
\bibliography{bib/ICRA.bib}

\end{document}